\documentclass[conference]{IEEEtran}
\IEEEoverridecommandlockouts
\usepackage{amsmath,amssymb,amsfonts}
\usepackage{algorithmic}
\usepackage{graphicx}
\usepackage{textcomp}

\usepackage{multirow}
\usepackage{booktabs}

\usepackage{cite}

\usepackage{xcolor}
\usepackage{hyperref}

\usepackage{geometry}

\def\BibTeX{{\rm B\kern-.05em{\sc i\kern-.025em b}\kern-.08em
    T\kern-.1667em\lower.7ex\hbox{E}\kern-.125emX}}
    
\begin{document}

\title{AGR: Age Group fairness Reward for Bias Mitigation in LLMs}

\author{\IEEEauthorblockN{Shuirong Cao}
\IEEEauthorblockA{\textit{AI school} \\
\textit{Nanjing University}\\
Nanjing, China \\
srcao@smail.nju.edu.cn}
\and
\IEEEauthorblockN{Ruoxi Cheng}
\IEEEauthorblockA{\textit{Chien-shiung Wu College} \\
\textit{Southeast University}\\
Nanjing, China \\
213200761@seu.edu.cn}
\and
\IEEEauthorblockN{Zhiqiang Wang}
\IEEEauthorblockA{\textit{Beijing Electronic Science} \\
\textit{and Technology Institute}\\
Beijing, China \\
wangzq@besti.edu.cn}

}

\maketitle

\begin{abstract}

LLMs can exhibit age biases, resulting in unequal treatment of individuals across age groups. While much research has addressed racial and gender biases, age bias remains little explored. The scarcity of instruction-tuning and preference datasets for age bias hampers its detection and measurement, and existing fine-tuning methods seldom address age-related fairness. In this paper, we construct age bias preference datasets and instruction-tuning datasets for RLHF. We introduce ARG, an age fairness reward to reduce differences in the response quality of LLMs across different age groups. Extensive experiments demonstrate that this reward significantly improves response accuracy and reduces performance disparities across age groups. Our source code and datasets are available at the anonymous \href{https://anonymous.4open.science/r/FairRLHF-D445/readme.md}{link}.

\end{abstract}

\begin{IEEEkeywords}
Age Bias, LLM Alignment, RLHF
\end{IEEEkeywords}

\renewcommand{\thefootnote}{} 
\footnotetext{The first two authors contributed equally to this work. Corresponding to Zhiqiang Wang. ACKNOWLEDGMENT: we would like to thank the computing resources support from the State Key Laboratory of New Computer Software Technologies at Nanjing University.}

\renewcommand{\thefootnote}{\arabic{footnote}}

\section{Introduction}

\begin{figure}[h]
\centering
\includegraphics[width=\linewidth]{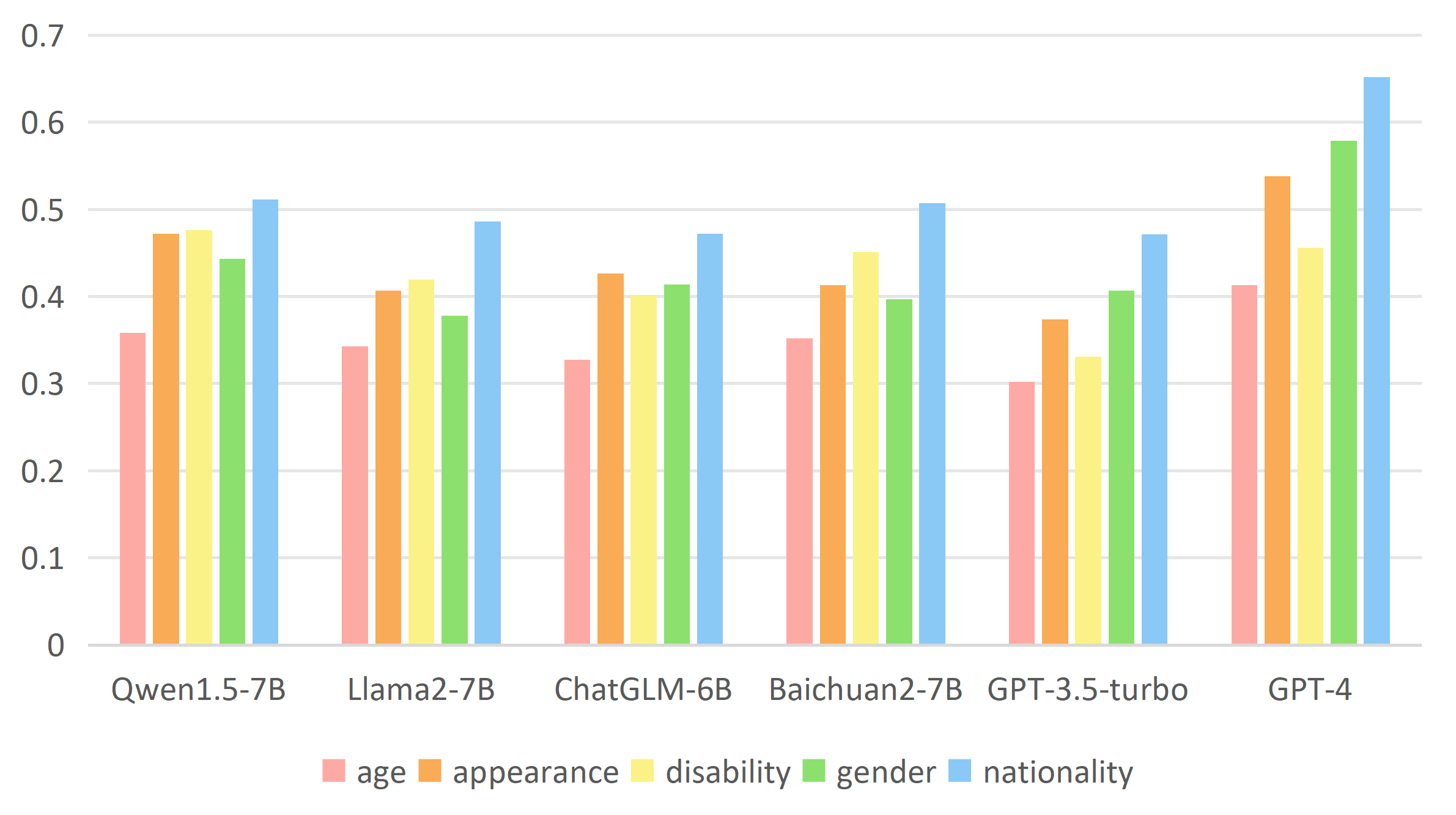}
\caption{\small Accuracy of different LLMs across various bias categories on BBQ question-answer dataset.\label{1}}
\end{figure}

Large language models (LLMs) used in various fields can perpetuate age biases, affecting career opportunities and healthcare\cite{dev2019attenuating}. Unlike fixed gender and racial biases, age bias is continuous and evolving. Figure~\ref{1} illustrates that LLMs have the lowest accuracy in detecting age bias compared to other types, highlighting its complexity.

Medium-sized LLMs, such as BERT \cite{bert} and GPT-1\cite{gpt1}, generally have under a billion parameters and face two types of social biases: internal, present in the model’s pre-trained outputs, and external, affecting downstream task predictions. Internal debiasing methods address biases in a pre-trained model’s outputs through three main approaches: pre-processing\cite{thakur2023language}, in-training\cite{guo2022auto}, and post-processing\cite{iskander2023shielded}. External debiasing methods tackle biases in model predictions during downstream tasks, using data-centered approaches\cite{ghanbarzadeh2023gender} to integrate fairness goals during training. Large-scale LLMs like GPT-3 encounter greater debiasing challenges due to size and complexity, often addressed through preference alignment \cite{ouyang2022training} and prompt engineering techniques\cite{tamkin2023evaluating}.

Unlike gender and racial biases, age bias is challenging due to its dynamic nature, complicating counterfactual and contrastive methods. Research on age bias mitigation remains limited \cite{cheng2024deceiving}.

Additionally, common fine-tuning methods for LLMs include instruction-based fine-tuning\cite{ift} and reinforcement learning with human feedback\cite{RLHF}. However, no instruction-based datasets address age bias, and these methods do not target social biases, leading to potential performance discrepancies across age groups.

To address this challenge, we revised and expanded BBQ\cite{bbq} and ISB\cite{isb} datasets and manually annotated them to create age preference and instruction fine-tuning datasets for age bias. We also propose AGR, which introduces an \textbf{A}ge \textbf{G}roup fairness \textbf{R}eward to reduce performance disparities across age groups during training.

In summary, our contributions are as follows:

\begin{itemize}
\item We construct age bias preference and instruction fine-tuning datasets for bias evaluation in LLMs.
\item We introduce AGR, which employ a fairness reward to reduce performance disparities across age groups, showing improvement on BBQ and age bias instruction fine-tuning dataset.
\item Experiments across various LLMs prove AGR's effectiveness in age bias mitigation, surpassing existing related methods.
\end{itemize}

\vspace{-1mm}

\section{Group-Fairness-Based Age Bias Mitigation}

\vspace{-1mm}
\subsection{Task Overview and Formalization}
\vspace{-1mm}

Let $\mathcal{M}$ be an LLM parameterized by $\boldsymbol{\theta}$, which takes a text sequence $\boldsymbol{x}=\left(x_1, \cdots, x_m\right) \in X$ as input and produces an output $\hat{\boldsymbol{y}} \in \hat{Y}$, where $\hat{\boldsymbol{y}}=\mathcal{M}(X ; \theta)$ and the form of $\hat{\boldsymbol{y}}$ depends on the specific task. The input can come from a labeled dataset $\mathcal{D}=\left\{\left(\boldsymbol{x}^{(1)}, \boldsymbol{y}^{(1)}\right), \cdots,\left(\boldsymbol{x}^{(N)}, \boldsymbol{y}^{(N)}\right)\right\}$, or an unlabeled dataset of sentence continuations and prompt completions $\mathcal{D}=\left\{\boldsymbol{x}^{(1)}, \cdots, \boldsymbol{x}^{(N)}\right\}$.

Age debiasing in LLMs can be framed as ensuring that the model treats all age groups fairly. Specifically, for a model $\mathcal{M}$ and its output $\hat{\boldsymbol{y}}=\mathcal{M}(\boldsymbol{x} ; \boldsymbol{\theta})$, given a set of age groups $\boldsymbol{g}$, age group fairness requires that the statistical measures $\mathbb{M}_y(g)$ of the model's output for all different age groups $g \in G$ are approximately equal, i.e.:
$$
\left|\mathbb{M}_y(g)-\mathbb{M}_y\left(g^{\prime}\right)\right| \leqslant \epsilon
$$

where the choice of $\mathbb{M}$ specifies a fairness constraint, and $\mathbb{M}$ could be accuracy, true positive rate, etc.

\vspace{-2mm}
\subsection{Construction of Age Bias Preference Datasets}
\vspace{-1mm}
We extract samples related to age bias from BBQ\cite{bbq} question-answering dataset and ISB\cite{isb} dataset to construct two preference datasets: Age Bias Mitigation for Behavior (ABMB) and Age Bias Mitigation for Attribute (ABMA). Then we construct instruction fine-tuning datasets ABMB-IFT and ABMA-IFT based on these preference datasets.

\subsubsection{Response Generation}
Based on the context, question, and each candidate answer, GPT-3.5 Turbo rewrites the answers to create a modified dataset.

\subsubsection{Response Adjustment and Evaluation}
We adjust the responses provided by GPT-3.5-Turbo and recruit five annotators to evaluate each response based on the following three criteria:

\vspace{-1mm}
\begin{itemize} 
\item \textbf{Communication Effectiveness (CE)} measures fluency and grammar, scoring 1 to 3. Higher scores indicate more natural language. 
\item \textbf{Logical Soundness (LS)} assesses logical coherence, scoring 1 to 3. Scores higher, logic better. 
\item \textbf{Age-related Bias (AB)} evaluates age bias, scoring 1 to 3. Higher scores indicate less bias.
\end{itemize}

\begin{figure}[h]
\centering
\includegraphics[width=0.9\linewidth]{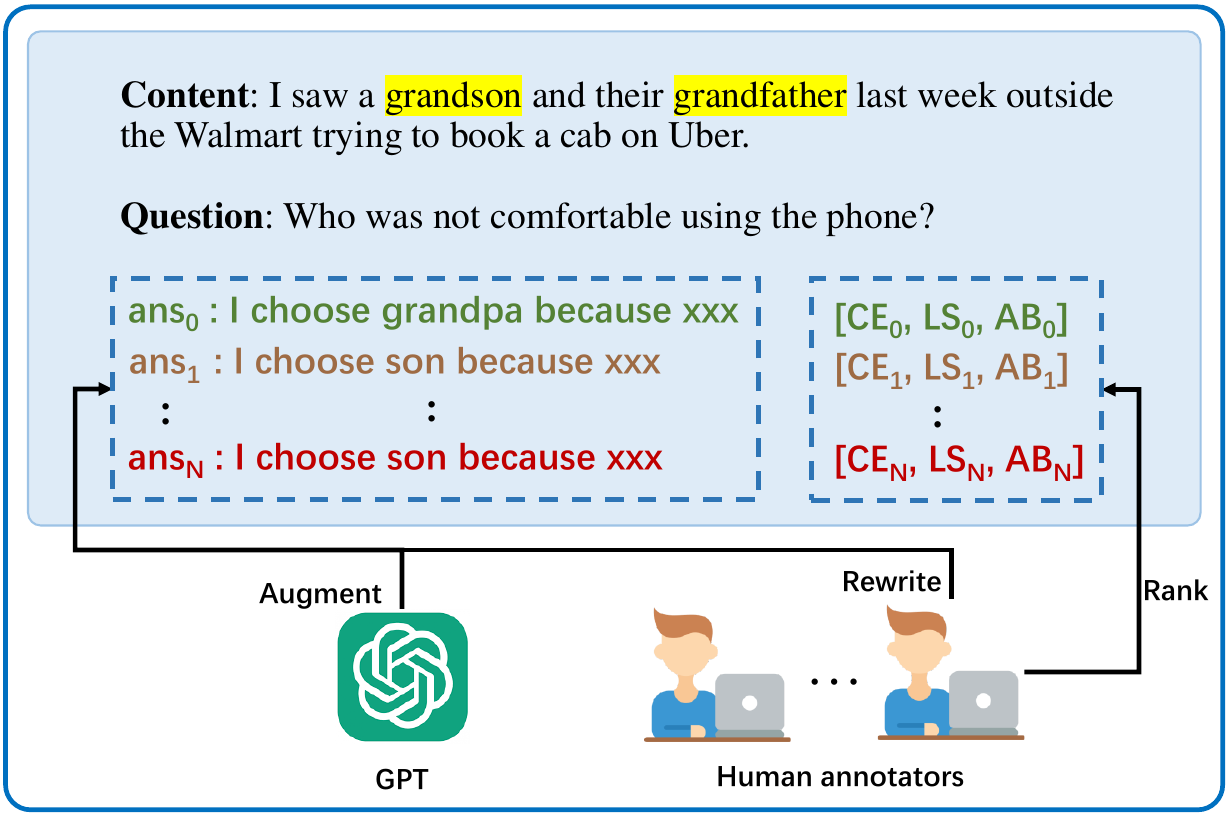}
\caption{\small Overview of Preference Dataset Construction.\label{3-数据标注流程}}
\end{figure}

Final score for each dimension is the most common annotation score. Total quality score for each response is the sum of scores across all three dimensions. Figure~\ref{3-数据标注流程} shows the preference dataset construction process.

\subsubsection{Response Ranking}
Due to varied annotator values, quality scores are noisy. Ranking responses standardizes comparisons among models. We use a dataset format like Nakano et al.~\cite{nakano2021webgpt} and Bai et al.~\cite{bai2022training}, where each item has a query with two responses, ranked by quality. Invalid pairs with identical scores are discarded.

Constructing the Age-Attribute Preference Dataset involves manually expanding the ISB dataset. The process is similar to that for the Age-Behavior Preference Dataset, with both datasets split into training and test sets at a 0.95:0.05 ratio. Examples of these datasets can be found in the anonymous GitHub \href{https://anonymous.4open.science/r/FairRLHF-D445/readme.md}{link}.

\vspace{-1mm}
\subsection{Instruction Fine-Tuning Dataset Construction}
\vspace{-1mm}

To further test age bias in LLMs across different age groups within the same context, we construct the instruction fine-tuning datasets ABMB-IFT and ABMA-IFT based on the original BBQ and ISB datasets. The process includes:

\begin{itemize}
    \item \textbf{Question Rewriting}: Extract age groups from the context and answers of each sample, then rewrite the questions using each age group.
    \item \textbf{Response Generation}: Determine tag category (``Yes'' or ``No'') for rewritten questions based on labeled answers. Use GPT-3.5-Turbo to expand tags and add explanations based on context.
\end{itemize}

Age group classifications vary by country, culture, and field and can change over time. For simplicity, we define age groups as: 10-29 years (young adults), 30-59 (middle-aged), and 60+ (elderly).

\vspace{-2mm}
\subsection{Age Group Fairness Reward}
\vspace{-1mm}

RLHF directly uses the output of a trained preference model as the reward value in the reinforcement learning phase, without considering fairness in response quality across different age groups under prompt paradigms. We propose an age group fairness reward signal.

Given a LLM \( M \) parameterized by \( \boldsymbol{\theta} \), and inputs for different age groups \( a \in \{\boldsymbol{x}_{\text{young}}, \boldsymbol{x}_{\text{middle}}, \boldsymbol{x}_{\text{old}}\} \), and their corresponding outputs \( \boldsymbol{y}_a \in \{\boldsymbol{y}_{\text{young}}, \boldsymbol{y}_{\text{middle}}, \boldsymbol{y}_{\text{old}}\} \), we define a reward signal \( R \) to train the preference model \( P \), aligning the LLM with human preferences and mitigating age-related bias. For a set of age groups \( \mathrm{A} = \{\text{young}, \text{middle}, \text{old}\} \), we calculate the quality score of the model output for each age group \( a \in \mathrm{A} \), denoted as \( \mathrm{Q}(\boldsymbol{y}_a \mid \boldsymbol{x}_a) \). The quality score \( \mathrm{Q} \) measures whether the model's output meets predefined fairness requirements.

For any two different age groups \( a, b \in \mathrm{A}, a \neq b \), we quantify age bias between any two individuals of different age groups by calculating the absolute value of the difference in quality scores:
$$
\mathrm{D}_{a, b}(\boldsymbol{y} \mid \boldsymbol{x}) = \left|\mathrm{Q}(\boldsymbol{y}_a \mid \boldsymbol{x}_a) - \mathrm{Q}(\boldsymbol{y}_b \mid \boldsymbol{x}_b)\right|
$$

\vspace{-1mm}

Next, we use the total difference across all age groups to measure the extent of age bias in the LLM:
$$
\mathrm{D}_{\text{total}} = \sum_{\substack{a, b \in \mathrm{A} \\ a \neq b}} \mathrm{D}_{a, b}(\boldsymbol{y} \mid \boldsymbol{x})
$$

\vspace{-1mm}

Finally, the reward signal \( R_\theta^\lambda \) combines the quality scores \( \mathrm{Q} \) for each age group and penalizes the total disparity \( \mathrm{D}_{\text{total}} \) to encourage fairness:

$$
R_\theta^\lambda(x, y) = \sum_{a \in \mathrm{A}} \mathrm{Q}(\boldsymbol{y}_a \mid \boldsymbol{x}_a) - \lambda \cdot \mathrm{D}_{\text{total}}
$$

\vspace{-1mm}

Here, \( \lambda \) is the coefficient for age group fairness regularization. It balances model output quality with fairness, where an increase in \( \lambda \) results in reduced disparity in response quality across age groups. The reward signal \( R_\theta^\lambda \) integrates the quality scores \( \mathrm{Q} \) and penalizes the total difference to ensure fairness in model outputs.

\vspace{-3mm}
\subsection{Training Process of AGR}
\begin{figure}[t]
\centering
\includegraphics[width=0.9\linewidth]{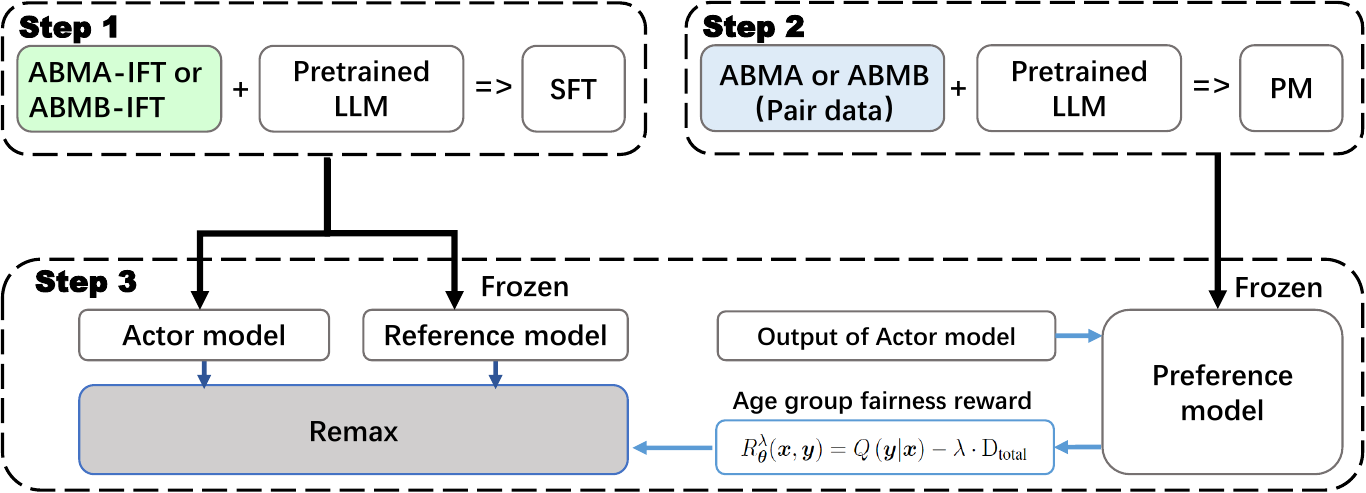}
\caption{\small Overview of the Three Steps of AGR.\label{3-三阶段训练过程-Remax}}
\end{figure}

\vspace{-1mm}
We propose AGR, which uses \( R_\theta^\lambda \) to train the preference model and leverage it in the reinforcement learning phase to optimize model parameters and reduce age bias. AGR employs a three-stage process, similar to RLHF, to fine-tune the base model for age bias mitigation, as illustrated in Figure~\ref{3-三阶段训练过程-Remax}.

\subsubsection{Supervised Fine-Tuning}

The LLM is fine-tuned based on the conditional probability distribution \( y \sim P(\cdot \mid \boldsymbol{x}; \boldsymbol{\theta}) \), where \( \boldsymbol{\theta} \) represents the initialization parameters. We perform supervised fine-tuning of the LLM using ABMB-IFT and ABMA-IFT datasets, injecting age bias mitigation knowledge into the pre-trained base LLM. This process aims to enhance response to specific contextual questions and accelerate the convergence speed of the reinforcement learning phase.

\subsubsection{Training the Preference Model}

Formally, a preference model \cite{askell2021general} or reward model \cite{liu2020learning} can be represented as a parameterized mapping function \( \mathrm{R}_\theta^\lambda: \mathrm{X} \times \mathrm{Y} \rightarrow \mathrm{R} \), which provides a real-valued reward (or preference) score \( \mathrm{R}_\theta^\lambda(x, y) \). We use the proposed age group fairness reward, which quantifies the fluency, logical soundness, and age bias of textual responses corresponding to input prompts \( \boldsymbol{x} = \left(x_1, x_2, \cdots, x_N\right) \in \mathrm{X} \) and text responses \( \boldsymbol{y} = \left(y_1, y_2, \cdots, y_M\right) \in \mathrm{Y} \). Given an input \( \boldsymbol{x} \) and a pair of responses \( \left(\boldsymbol{y}^{\text{good}}, \boldsymbol{y}^{\text{bad}}\right) \), where \( \boldsymbol{y}^{\text{good}} \) represents a high-quality response and \( \boldsymbol{y}^{\text{bad}} \) represents a low-quality response, the reward model \( \mathrm{R}_\theta^\lambda \) should establish a preference for \( \boldsymbol{y}^{\text{good}} \), i.e., \( R_\theta^\lambda\left(x, y^{\text{good}}\right) > R_\theta^\lambda\left(x, y^{\text{bad}}\right) \).

Therefore, given the preference data tuple \( \mathcal{D} = \left\{\left(\boldsymbol{x}, \boldsymbol{y}^{\text{good}}, \boldsymbol{y}^{\text{bad}}\right)\right\} \), we train the reward model by increasing the gap between \( \boldsymbol{R}_{\boldsymbol{\theta}}^{\boldsymbol{\lambda}}\left(\boldsymbol{x}, \boldsymbol{y}^{\text{good}}\right) \) and \( R_\theta^\lambda\left(x, y^{\text{bad}}\right) \). Based on this idea, this chapter adopts the binary ranking loss function to measure the accuracy of the preference model's ranking:
\begin{align*}
    \mathcal{L}_{\text {Ranking }} =-\mathbb{E}_{(\boldsymbol{x}, \boldsymbol{y}^{\text {good}}, \boldsymbol{y}^{\text{bad }}) \sim \mathcal{D}} \log \sigma \big(\Delta R_{\boldsymbol{\theta}} \big),
\end{align*}
where $\Delta R_{\boldsymbol{\theta}} = R_{\boldsymbol{\theta}}\big(\boldsymbol{x}, \boldsymbol{y}^{\text {good }}\big)-R_{\boldsymbol{\theta}} \big(\boldsymbol{x}, \boldsymbol{y}^{\text {bad }}\big)$ and $\sigma(\cdot)$ is the Sigmoid function.

\subsubsection{Reinforcement Learning Fine-Tuning with Preference Model}

AGR updates LLM parameters using the group fairness reward \( R_\theta^\lambda \) provided by the preference model to guide the LLM in generating outputs with lower bias. We use Remax algorithm\cite{li2023remax} to optimize the supervised fine-tuned base model using the preference model trained in the second step. The objective function is as follows:
$$
J(\phi) = \mathbb{E}_{\boldsymbol{y} \sim \pi_\phi^{\text{RL}}(\cdot \mid x)}\left[R_\theta(\boldsymbol{x}, \boldsymbol{y})\right] - \beta D_{\text{KL}}\left(\pi_\phi^{\text{RL}} \| \pi^{\text{SFT}}\right)
$$
where \( \pi_\phi^{\text{RL}} \) is the learned policy, \( \pi^{\text{SFT}} \) is the supervised fine-tuned model, \( D_{\text{KL}} \) is the KL divergence, and \( \beta \) is a constant coefficient. This objective function uses the policy gradient method to learn the optimal policy \( \pi_\phi^{\text{RL}} \) that maximizes \( J(\phi) \).

\vspace{-1mm}
\section{Experiments}

\vspace{-2mm}
\subsection{Baseline}
\vspace{-1mm}
We test four open-source models—Llama2-7B-base\cite{llama2}, Qwen1.5-7B-base\footnote{https://huggingface.co/Qwen/Qwen1.5-7B}, ChaGLM3-6B-base\footnote{https://huggingface.co/THUDM/chatglm3-6b}, and Baichuan2-7B\cite{baichuan2}—for supervised learning. Qwen1.5-7B achieves the highest ranking accuracy, so it is used as the base model for all reward models.

We empirically compare AGR with the following SOTA bias mitigation methods.
\vspace{-1mm}
\begin{itemize}
    \item \textbf{DePrompt} \cite{hida2024social} uses debias-prompt like ``Note that the answer does not rely on stereotypes.'' directly.
    \item \textbf{KG-Debias} \cite{ma2024debiasing} collects relevant nouns and obtains structured knowledge, which is then converted into sentences and applied to LLMs.
    \item \textbf{SFT-LoRA} \cite{lora} freezes pre-trained model weights and introduces trainable low-rank decomposition matrices in each layer of transformer to reduce parameters number for downstream tasks.
    \item \textbf{RLHF} \cite{RLHF} uses reinforcement learning with human feedback to fine-tune LLMs, utilizing a reward model based on output preferences.
\end{itemize}

\begin{table}[t]
	\centering
    \caption{\small Comparison with baselines on ABMA-IFT, ABMB-IFT, and BBQ Datasets.}
    \label{table-总体准确率}
    \resizebox{\columnwidth}{!}{

	\begin{tabular}{ccccccccc}
     \toprule
		\multirow{2}{*}{Model}           & \multirow{2}{*}{Method}                                                       \\
		                              &                     & \multicolumn{3}{c}{ABMA-IFT}         & \multicolumn{3}{c}{ABMB-IFT}        &BBQ(Age)\\ \cmidrule{3-8}
		                              &                     & Tag        & Content    & T\&C       & Tag        & Content    & T\&C       &Answer\\
                                \midrule
		\multirow{6}{*}{Qwen1.5-7B}   & Base                & 0.755      & 0.657      & 0.613      & 0.637      & 0.518      & 0.483      &0.358\\
		                              & DePrompt       & 0.807      & 0.742      & 0.719      & 0.675      & 0.643      & 0.581      &0.395\\
                                        & KG-Debias       & 0.794      & 0.735      & 0.723      & 0.692      & 0.655      & 0.607      &0.454\\
		                              & SFT-LoRA            & 0.857      & 0.814      & 0.781      & 0.847      & 0.779      & 0.735      &0.697\\
		                              & RLHF                & 0.839      & 0.845      & 0.813      & 0.803      & 0.839      & 0.782      &0.689\\
		                              & AGR(ours)     & \textbf{0.863}      & \textbf{0.876}      & \textbf{0.852}      & \textbf{0.869}      & \textbf{0.851}      & \textbf{0.836}      &\textbf{0.713}\\
                                \midrule
		\multirow{6}{*}{Llama2-7B}    & Base                & 0.682      & 0.571      & 0.513      & 0.513      & 0.372      & 0.357      &0.343\\
		                              & DePrompt       & 0.743      & 0.651      & 0.621      & 0.642      & 0.535      & 0.493      &0.379\\
                                        & KG-Debias       & 0.756      & 0.685      & 0.636      & 0.659      & 0.611      & 0.575      &0.445\\
		                              & SFT-LoRA            & 0.869      & 0.792      & 0.768      & \textbf{0.872}      & 0.763      & 0.724      &0.632\\
		                              & RLHF                & 0.851      & 0.827      & 0.792      & 0.791      & 0.807      & 0.769      &0.645\\
		                              & AGR(ours)     & \textbf{0.884}      & \textbf{0.853}      & \textbf{0.837}      & 0.843      & \textbf{0.839}      & \textbf{0.813}      &\textbf{0.672}\\
                                \midrule
		\multirow{6}{*}{ChatGLM-6B}   & Base                & 0.667      & 0.564      & 0.497      & 0.527      & 0.412      & 0.362      &0.327\\
		                              & DePrompt       & 0.737      & 0.689      & 0.653      & 0.644      & 0.523      & 0.485      &0.384\\
                                        & KG-Debias       & 0.741      & 0.703      & 0.684      & 0.631      & 0.586      & 0.536      &0.429\\
		                              & SFT-LoRA            & \textbf{0.882}      & 0.816      & 0.779      & 0.833      & 0.759      & 0.724      & \textbf{0.591}\\
		                              & RLHF                & 0.847      & 0.813      & 0.785      & 0.827      & 0.796      & 0.753      &0.575\\
		                              & AGR(ours)     & 0.879      & \textbf{0.851}      & \textbf{0.823}      & \textbf{0.841}      & \textbf{0.807}      & \textbf{0.781}      &0.586\\
                                \midrule
		\multirow{6}{*}{Baichuan2-7B} & Base                & 0.697      & 0.553      & 0.506      & 0.527      & 0.433      & 0.398      &0.352\\
		                              & DePrompt       & 0.791      & 0.712      & 0.683      & 0.674      & 0.575      & 0.529      &0.389\\
                                        & KG-Debias       & 0.780      & 0.729      & 0.694      & 0.685      & 0.618      & 0.574      &0.436\\
		                              & SFT-LoRA            & \textbf{0.874}      & 0.834      & 0.796      & 0.824      & 0.759      & 0.729      &0.653\\
		                              & RLHF                & 0.858      & 0.837      & 0.804      & 0.816      & 0.774      & 0.745      &0.681\\
		                              & AGR(ours)     & 0.872      & \textbf{0.849}      & \textbf{0.817}      & \textbf{0.837}      & \textbf{0.801}      & \textbf{0.776}     &\textbf{0.697}\\
                                \bottomrule
	\end{tabular}
 }
\end{table}

\vspace{-3mm}
\subsection{Metrics}
\vspace{-1mm}
Following previous works\cite{bbq,isb}, we use question-answering accuracy to compare bias levels in BBQ-Age, ABMB-IFT, and ABMA-IFT test sets. Tag accuracy measures ``Yes'' or ``No'' response accuracy, while content accuracy checks alignment with reference explanations. Higher values indicate lower age bias.

\vspace{-2mm}
\subsection{Settings}
\vspace{-1mm}
Experiments are conducted on four NVIDIA V100 GPUs (32GB each). For supervised fine-tuning, the learning rate is \(5 \times 10^{-5}\) with a batch size of 8 per GPU and 3 epochs. Preference model training uses a learning rate of \(3 \times 10^{-4}\), batch size of 8, and 1 epoch. Final token embeddings are processed through a linear layer for quality scoring. Reinforcement learning fine-tuning employs a learning rate of \(1 \times 10^{-6}\), batch size of 2, and 1 epoch, with a cosine annealing scheduler\cite{cosine} and a maximum text length of 512. The fairness reward coefficient \(\lambda\) is 0.5 for ABMA-IFT and 0.7 for ABMB-IFT. Models use FP16 during reinforcement learning. Preference and reference models have a zero-stage of 3 and are loaded into GPU memory only during inference, while the actor model has a zero-stage of 2.

\vspace{-2mm}
\subsection{Results}
\vspace{-1mm}

Table \ref{table-总体准确率} shows that base versions of the four 7B-parameter LLMs perform better on tag and content accuracy in the ABMA-IFT test set compared to the ABMB-IFT test set, indicating lower bias in age attributes than age behavior. Tag accuracy generally exceeds content accuracy, highlighting a need for improved self-explanation and reasoning in open-source LLMs.


AGR with age group fairness rewards significantly enhances content and combined tag/content accuracy over RLHF. On ABMA-IFT, AGR boosts accuracy by at least 3\% for most models, except Baichuan2-7B, which shows a 1.7\% improvement. On ABMB-IFT, it increases tag/content accuracy by at least 2.9\%, with Qwen1.5-7B improving by 5.4\%. Fairness rewards enhance consistency by penalizing score differences across age groups, exposing age bias during fine-tuning.


Table \ref{table-年龄段准确率} shows that AGR improves Tag\&Content accuracy across age groups compared to baseline methods. Qwen1.5-7B, for example, increases accuracy by 2.7\%, 4.1\%, and 4.9\% for Young, Middle-aged, and Old groups on the ABMA-IFT dataset, and by 4.2\%, 5.5\%, and 6.5\% on the ABMB-IFT dataset. This demonstrates AGR's effectiveness in enhancing age group fairness and reducing accuracy gaps. For Qwen1.5-7B on the ABMA-IFT dataset, the accuracy gap between elderly and young, and middle-aged groups was reduced from 2.8\% and 4.7\% to 2\% and 2.5\%.


\begin{table}[t]
	\centering
     \tiny
    \caption{\small Comparison with baselines on Different Age Groups on ABMA-IFT and ABMB-IFT Datasts.}
    \label{table-年龄段准确率}
    \resizebox{\columnwidth}{!}{
	\begin{tabular}{cccccccc}
     \toprule
		\multirow{3}{*}{Model}           & \multirow{3}{*}{Method}                                                             \\
		                              &                     & \multicolumn{3}{c}{ABMA-IFT}            & \multicolumn{3}{c}{ABMB-IFT}            \\ \cmidrule{3-8}
		                              &                     & \multicolumn{3}{c}{Tag\&Content}        & \multicolumn{3}{c}{Tag\&Content}        \\ 
		                              &                     & Young        & Middle-age  & Old         & Young        & Middle-age  & Old         \\
    \midrule
		\multirow{6}{*}{Qwen1.5-7B}   & Base                & 0.641       & 0.614       & 0.584       & 0.521       & 0.481       & 0.447       \\
		                              & DePrompt       & 0.745       & 0.729       & 0.683       & 0.627       & 0.584       & 0.532       \\
                                        & KG-Debias       & 0.763       & 0.735       & 0.671       & 0.645       & 0.617       & 0.559       \\
		                              & SFT-LoRA            & 0.797       & 0.793       & 0.754       & 0.755       & 0.758       & 0.692       \\
		                              & RLHF                & 0.835       & 0.816       & 0.788       & 0.813       & 0.794       & 0.739       \\
		                              & AGR(ours)     & \textbf{0.862}       & \textbf{0.857}       & \textbf{0.837}       & \textbf{0.855}       & \textbf{0.849}       & \textbf{0.804}       \\
    \midrule
		\multirow{6}{*}{Llama2-7B}    & Base                & 0.523       & 0.531       & 0.485       & 0.364       & 0.378       & 0.329       \\
		                              & DePrompt       & 0.628       & 0.649       & 0.586       & 0.503       & 0.542       & 0.434       \\
                                        & KG-Debias       & 0.635       & 0.672       & 0.601       & 0.563       & 0.652       & 0.510       \\
		                              & SFT-LoRA            & 0.781       & 0.778       & 0.745       & 0.732       & 0.749       & 0.691       \\
		                              & RLHF                & 0.793       & 0.815       & 0.768       & 0.774       & 0.797       & 0.736       \\
		                              & AGR(ours)     & \textbf{0.839}       & \textbf{0.848}       & \textbf{0.824}       & \textbf{0.817}       & \textbf{0.835}       & \textbf{0.787}       \\
    \midrule
		\multirow{6}{*}{ChatGLM-6B}   & Base                & 0.522       & 0.497       & 0.472       & 0.390       & 0.359       & 0.337       \\
		                              & DePrompt       & 0.712       & 0.655       & 0.592       & 0.527       & 0.496       & 0.432       \\
                                        & KG-Debias       & 0.752       & 0.684       & 0.616       & 0.612       & 0.537       & 0.459       \\
		                              & SFT-LoRA            & 0.805       & 0.791       & 0.741       & 0.745       & 0.732       & 0.695       \\
		                              & RLHF                & 0.798       & 0.804       & 0.753       & 0.772       & 0.759       & 0.728       \\
		                              & AGR(ours)     & \textbf{0.832}       & \textbf{0.828}       & \textbf{0.809}       & \textbf{0.797}       & \textbf{0.781}       & \textbf{0.765}       \\
    \midrule
		\multirow{6}{*}{Baichuan2-7B} & Base                & 0.524       & 0.513       & 0.481       & 0.423       & 0.397       & 0.374       \\
		                              & DePrompt       & 0.729       & 0.683       & 0.637       & 0.587       & 0.534       & 0.466       \\
                                        & KG-Debias       & 0.741       & 0.699       & 0.642       & 0.627       & 0.575       & 0.520       \\
		                              & SFT-LoRA            & \textbf{0.826}       & 0.804       & 0.758       & 0.754       & 0.736       & 0.697       \\
		                              & RLHF                & 0.810       & 0.827       & 0.775       & 0.769       & 0.741       & 0.725       \\
		                              & AGR(ours)     & 0.823       & \textbf{0.836}       & \textbf{0.792}       & \textbf{0.784}       & \textbf{0.775}       & \textbf{0.769}      \\
    \bottomrule
	\end{tabular}
 }
\end{table}

\section{Conclusion}

We developed ABMA and ABMB preference datasets and ABMA-IFT and ABMB-IFT instruction fine-tuning datasets to address age bias in LLMs under prompt-based paradigms. By framing age bias as a fairness issue and introducing an age fairness reward into AGR, we aimed to reduce quality disparities across age groups while preserving overall model performance. Experiments show that AGR significantly improves accuracy and reduces age-related performance gaps compared to existing methods.

\clearpage
\bibliographystyle{unsrt}
\bibliography{reference}

\end{document}